\documentclass[sigconf,screen,nonacm]{acmart} 
\usepackage{booktabs} 

\usepackage{amsmath,amsfonts,bm}

\def\eqref#1{equation~\ref{#1}}

\def\1{\bm{1}}

\newcommand{\bs}{\boldsymbol}
\newcommand{\ba}{\bs{a}}
\newcommand{\bA}{\bs{A}}

\newcommand{\bD}{\bs{D}}
\newcommand{\bI}{\bs{I}}
\newcommand{\bS}{\bs{S}}
\newcommand{\bW}{\bs{W}}
\newcommand{\bx}{\bs{x}}
\newcommand{\bX}{\bs{X}}
\newcommand{\by}{\bs{y}}
\newcommand{\bY}{\bs{Y}}

\newcommand{\balpha}{\bs{\alpha}}
\newcommand{\btheta}{\bs{\theta}}

\DeclareMathAlphabet{\mathsfit}{\encodingdefault}{\sfdefault}{m}{sl}
\SetMathAlphabet{\mathsfit}{bold}{\encodingdefault}{\sfdefault}{bx}{n}

\setcopyright{rightsretained}

\acmConference[DLG'21]{The 6th International Workshop on Deep Learning on Graphs: Methods and Applications}{August 14 - 18, 2021}{Virtual Conference}
\acmYear{2021}
\copyrightyear{2021}

\acmPrice{}
\acmISBN{}
\acmDOI{}

\usepackage{amsmath}
\usepackage{amscd,amsfonts,amsbsy,rotating}
\usepackage{balance}
\usepackage{graphicx}
\usepackage{epsfig,epstopdf}
\usepackage{subfigure}
\usepackage{multirow}
\usepackage{booktabs}
\usepackage{color,xcolor}
\usepackage{url}
\usepackage{latexsym,bm}
\usepackage{enumitem,balance,mathtools}
\usepackage{wrapfig}
\usepackage{euscript}
\usepackage{ifpdf}
\usepackage{diagbox}
\usepackage{caption}
\usepackage{makecell}
\usepackage{subfigure}
\usepackage[leftcaption]{sidecap}
\usepackage{bm}
\usepackage{textcomp}
\usepackage{upgreek}
\usepackage[normalem]{ulem}
\usepackage{breakurl}

\usepackage{array}
\newcolumntype{N}{@{}m{0pt}@{}}

\usepackage{lipsum}
\PassOptionsToPackage{hyphens}{url}\usepackage{hyperref}

\usepackage[ruled,linesnumbered]{algorithm2e}

\usepackage{algorithmic}

\newcommand{\minisection}[1]{\vspace{5pt}\noindent\textbf{#1.}}

\newcommand{\lossname}{{Loge }}
\newcommand{\lossshortname}{{loge }}

\begin{document}
\sloppy
  
\title{Bag of Tricks for Node Classification with\\Graph Neural Networks}
\author{Yangkun Wang$^{1\dag}$, Jiarui Jin$^{1\dag}$, Weinan Zhang$^{1\ddagger}$, Yong Yu$^1$, Zheng Zhang$^2$, David Wipf$^{2\ddagger}$}
\affiliation{$^1$Shanghai Jiao Tong University, $^2$Amazon}
\email{espylapiza@gmail.com, {jinjiarui97, wnzhang, yyu}@sjtu.edu.cn, {zhaz, daviwipf}@amazon.com}
\renewcommand{\shorttitle}{Bag of Tricks for Node Classification with Graph Neural Networks}
\renewcommand{\shortauthors}{Y. Wang, et al.}
\settopmatter{printacmref=false}

\begin{abstract}
Over the past few years, graph neural networks (GNN) and label propagation-based methods have made significant progress in addressing node classification tasks on graphs.  However, in addition to their reliance on elaborate architectures and algorithms, there are several key technical details that are frequently overlooked, and yet nonetheless can play a vital role in achieving satisfactory performance.  In this paper, we first summarize a series of existing tricks-of-the-trade, and then propose several new ones related to label usage,\footnote{For the label trick in particular, please see our more detailed, follow-up analysis \cite{wang2021why}.} loss function formulation, and model design that can significantly improve various GNN architectures.  We empirically evaluate their impact on final node classification accuracy by conducting ablation studies and demonstrate consistently-improved performance, often to an extent that outweighs the gains from more dramatic changes in the underlying GNN architecture.  Notably, many of the top-ranked models on the Open Graph Benchmark (OGB) leaderboard and KDDCUP 2021 Large-Scale Challenge MAG240M-LSC benefit from these techniques we initiated.
\end{abstract}

\maketitle
{
\renewcommand{\thefootnote}{\fnsymbol{footnote}}
\footnotetext[2]{Work done during internship at AWS Shanghai AI Lab.}
\footnotetext[3]{Corresponding authors.}
}

	
	\section{Introduction}
	Recently, machine learning tasks involving graphs have received increasing attention, among which node classification is one of the most prominent examples.
	Since the remarkable success of graph convolution networks (GCN) \citep{kipf2016semi}, many high-performance GNN designs have been proposed to address the node classification problem, such as graph attention networks (GAT) \citep{velivckovic2017graph} and GraphSAGE \citep{hamilton2017inductive}.
	At the same time, we have witnessed a steady improvement in model accuracy as demonstrated on the Open Graph Benchmark (OGB) leaderboard \citep{hu2020open}.
	For example, the top-$1$ test accuracy for node classification on the ogbn-arxiv dataset has improved from $70.1\%$ (based on node2vec \citep{grover2016node2vec}) to $74.1\%$ (based on GAT).
	
	However, these advances are not derived exclusively from the development of model architectures.
	Refinements including data processing, loss function design and negative sampling also play a major role.
	Specifically, for semi-supervised learning, it is worthwhile to explore the effective use of the information contained in node features and/or labels.
	In this context, a common approach is to train GNN models that make predictions based on node features and model parameters; however, this strategy cannot directly utilize existing label information (beyond their influence on model parameters through training). 
	In contrast, label propagation algorithms (LPA) \citep{zhu2005semi} spread label information to make predictions, but cannot exploit node features.
	Although many recent attempts \citep{klicpera2018predict,huang2020combining} propose to integrate node features and label information by combining GNN and LPA, these approaches suffer from the inherent limitation that LPA requires neighboring nodes to share similar labels and cannot be applied to graphs with edge features.
	
	In this paper, we propose a series of novel techniques covering both label usage and architecture design.
    Specifically, we first develop a sampling technique that enables GNNs to leverage random subsets of \emph{original labels} as a model input.
    Based on this, we also design an iterative enhancement which utilizes the \emph{predicted labels} from the previous iteration as input for further training.
    Additionally, we propose a robust loss function and describe different variants of GAT designs.
	We evaluate these modifications and tricks on multiple GNN architectures and datasets, demonstrating that they often lead to significant improvement in node classification accuracy.  
	
	Notably, as of Jul. 2, 2021, all of the top 10 models on the ogbn-arxiv leaderboard, including AGDN \citep{sun2020adaptive}, C\&S \citep{huang2020combining}, FLAG \citep{kong2020flag} and UniMP \citep{shi2020masked}, have applied these methods or minor variations thereof.
	Moreover, on the more challenging ogbn-proteins dataset, we can obtain an ROC-AUC of 0.8765, which at the time of our post to the OGB leaderboard, outperformed all prior methods.
	And our label usage ideas in particular, which we were the first to propose for improving node classification,\footnote{Please see our original submission to the OGB leaderboard for ogbn-arxiv on Sept. 5, 2020 at \href{https://ogb.stanford.edu/docs/leader_nodeprop/}{https://ogb.stanford.edu/docs/leader\_nodeprop/} and the corresponding original code at \href{https://github.com/dmlc/dgl/tree/1a131f6b9db21b5183a65c3c75a6ed122345b616/examples/pytorch/ogb/ogbn-arxiv}{this github repo}.} have been followed by UniMP \citep{shi2020masked} among others, and have now been adopted in many submissions to the OGB leaderboard.  Overall, these techniques continue to be widely adopted, as an evidenced by the KDDCUP 2021 Large-Scale Challenge MAG240M-LSC \citep{hu2021ogb}, e.g., the released results\footnote{\url{https://ogb.stanford.edu/kddcup2021/results/}.} indicate that all of the top 3 approaches benefit from techniques we initiated. 

	\section{Background} \label{sec:background}
	Given a graph $G=(V,E)$, where $V=\{v_1,v_2,\ldots,v_N\}$ is the set of nodes and $E$ is the set of edges, we denote $\bA$ as the adjacency matrix and $\bD$ as the diagonal degree matrix.
	We assume that we have node features $\bX=(\bx_1,\ldots,\bx_N)^T$ and one-hot encoded label matrix $\bY=(\by_1,\ldots,\by_N)^T\in \mathbb R^{N\times C}$, with $C$ being the number of classes. Each node is associated with a feature vector $\bx_i$ and label $\by_i$, assuming that only the first $M$ nodes $\by_1,\by_2,\ldots,\by_M$ can be observed during training.
	For each dataset $\mathcal{D}=\{v_i,\bx_i,\by_i\}_{i=1}^N$ associated with a graph $G$, we have the training set $\mathcal{D}_{train}$ ($|\mathcal{D}_{train}|=M$) and the test set $\mathcal{D}_{test}$.
	The goal of the node classification task is to predict the labels of unlabeled nodes.
	Given the loss function $\ell(\hat \by_i,\by_i)$, the optimization objective is to minimize the aggregated cost $\mathcal{L}(\btheta)=\sum_{i=1}^M\ell(\hat \by_i,\by_i)$, where $\hat \by$ indicates the predicted label and $\btheta$ indicates model parameters.
	
	\minisection{Label Propagation Algorithm}
	LPA is a semi-supervised algorithm that predicts unlabeled nodes by propagating the observed labels across the edges of the graph, with the underlying assumption that two nodes connected by an edge in the graph are likely to have the same label.
	Letting $\bS=\bD^{-\frac12}\bA\bD^{-\frac12}$ be the symmetric normalized adjacency, LPA solves a linear system $\bY^*=(1-\lambda)(\bI-\lambda \bS)^{-1}\bY$ by iteratively computing $\bY^{(k+1)}=\lambda \bS\bY^{(k)}+(1-\lambda)\bY^{(0)}$, where $\bY^{(0)}$ is the label matrix of training nodes, padded with zeros for test nodes.
    While effective in many circumstances, LPA does not make use of node features as do the GNN models described next.
	
	
	\minisection{Graph Neural Networks}
	GNNs are a family of multi-layer feed-forward neural networks that transform and propagate layer-wise features across graph edges.
	Among these models, a GCN architecture is widely adopted, relying on the layer-wise propagation rule
	\begin{equation}
		\label{eqn:gcn}
		\bX^{(l+1)}=\sigma ( \bD^{-\frac12}\bA\bD^{-\frac12}\bX^{(l)}\bW^{(l)}),
	\end{equation}
	where $\bW^{(l)}$ denotes a trainable weight matrix of the $l$-th layer, $\sigma(\cdot)$ is an activation function, and $\bX^{(l)}$ represents the $l$-th layer node representations.
	GAT models further leverage masked self-attention layers to implicitly assign different weights to different neighboring nodes.
	Assuming $(v_i, v_j)\in E$ is an edge, then the layer-wise propagation rule of GAT is as follows:
    \begin{equation}
		\label{eqn:gat}
		\begin{gathered}
		\alpha_{ij}^{(l)}=\frac{\exp\left(\mathrm{LeakyReLU}\left(\ba^T[\bW^{(l)}\bx_i^{(l)}\parallel \bW^{(l)} \bx_j^{(l)}]\right)\right)}{\sum_{r\in\mathcal{N}(v_i)}\exp\left(\mathrm{LeakyReLU}\left(\ba^T[\bW^{(l)} \bx_i^{(l)}\parallel \bW^{(l)} \bx_r^{(l)}]\right)\right)},\\
		\bx^{(l+1)}_i=\sigma\left(\sum_{v_j\in \mathcal{N}(v_i)}\alpha^{(l)}_{ij}\bW^{(k)}\bx^{(l)}_j\right),
        \end{gathered}
	\end{equation}
	where $\ba$ is a trainable weight vector, $\mathcal{N}(v_i)$ denotes the neighbors of node $v_i$, and $\parallel$ represents the concatenation operation.  Note that unlike LPA, when inferring the labels of test nodes, GNN models do not make explicit use of the ground-truth labels of training nodes.
	
	\minisection{Combinations of Label and Feature Propagation}
	Since both LPA based on spreading observed labels and GNN architectures that propagate node features often achieve promising performance, it is worth exploring combinations thereof to potentially overcome their respective limitations.
    However, one of the major challenges is that simple combinations can lead to trivial degenerate solutions when these labels are provided as input to trainable models. 
    Although many recent attempts have been made to circumvent this problem, they still have various limitations.  For example, APPNP \citep{klicpera2018predict} does not actually propagate ground-truth training labels (only predicted labels), while C\&S \citep{huang2020combining} propagates ground-truth labels but only during inference; it is not trained end-to-end.
    Instead, we propose an approach in Section~\ref{sec:label_usage} that allows parallel propagation of node features and labels during both training and inference stages.
    Based on this, we further design a novel label reuse strategy on graphs, which propagates not only the true labels of training nodes but also the predicted labels of test nodes.
	
	\section{Existing Tricks}
    Among many useful strategies, here we briefly discuss sampling, data augmentation, renormalization, and residual connections, which can all be applied in various settings to improve performance.

	
	\minisection{Sampling}
	Sampling techniques \citep{chen2018fastgcn,zou2019layer,hamilton2017inductive} are often essential for the efficient training of GNNs.
    For example, recent methods such as FastGCN \citep{chen2018fastgcn} and LADIES \citep{zou2019layer} investigate layer-wise and layer-dependent importance sampling.
    Additionally, negative sampling methods \citep{mikolov2013distributed}, as first proposed to serve as a simplified version of noise contrastive estimation, can also play an important role, and are now widely adopted in web-scale graph mining approaches such as PinSAGE \citep{ying2018graph}.
	
	\minisection{Data Augmentation}
	For semi-supervised node classification tasks on graphs, over-fitting and over-smoothing \citep{li2018deeper} are two main obstacles in training GNNs.
	In order to surmount these obstacles, the DropEdge method \citep{rong2019dropedge} randomly removes a certain number of edges from the input graph, acting like a data augmenter and a message-passing reducer.
	In addition to modifying graph structures, another direction is inspired by the recent success of adopting adversarial training in computer vision \citep{xie2020adversarial} by adding gradient-based adversarial perturbations to the input features, while keeping graph structures unchanged \citep{kong2020flag}.
	
	\minisection{Renormalization}
	The renormalization trick was introduced in GCN models \citep{kipf2016semi} to alleviate the numerical instabilities and gradient explosion brought about by repeated application of Eq.~(\ref{eqn:gcn}) during training with many layers.  Specifically, we replace $\bI+\bD^{-\frac{1}{2}}\bA\bD^{-\frac{1}{2}}$ with $\tilde \bD^{-\frac{1}{2}} \tilde \bA \tilde \bD^{-\frac{1}{2}}$, where $\tilde \bA=\bA+\bI$ and $\tilde \bD=\bD+\bI$.
	
	\minisection{GCN with Residual Connections}
	A primitive form of GCN whose linear connection with different parameters added to the message passing formulation was introduced \citep{kipf2016semi}.
	Subsequently, there has also been a body of work using broader forms of residual connections \citep{rossi2020sign,li2020deepergcn}.
	One variant we find to be stable and robust adds a linear connection with free parameters to GCN with the renormalization trick:
	\begin{equation}
		\label{eqn:trickgcn}
		\bX^{(l+1)}=\sigma\left(\tilde \bD^{-\frac12}\tilde \bA\tilde \bD^{-\frac12}\bX^{(l)}\bW_0^{(l)}+\bX^{(l)}\bW_1^{(l)}\right).
	\end{equation}
	This form can avoid the gradient instabilities with proper initialization of $\bW$, and moreover makes the GCN more expressive and overcomes the over-smoothing issue, since the linear component in Eq.~(\ref{eqn:trickgcn}) retains the node representations distinguishable even with infinitely many propagation layers.
	
	\section{A New Bag of Tricks} \label{sec:bag_of_tricks}
	
	\subsection{Label Usage} \label{sec:label_usage}
	
	\minisection{Label as Input}
	For semi-supervised classification tasks, apart from the graph $G$ and the feature matrix $\bX$, we also have access to the label matrix $\bY$, in which some nodes have missing labels and need to be predicted.
	However, outside of LPA, prior work seldom considers the explicit use of ground-truth label information during the inference of test node labels.
	Instead, the label information is usually regarded only as the target for the supervised training of GNN models.
    However, when the training accuracy is below $100\%$, the label information of the misclassified samples is not contained in the model, despite the fact that they can provide additional information during inference.
	Additionally, samples misclassified by the model have the potential to mislead their neighbors.

	While some approaches are proposed to address this problem by combining GNN with LPA, they have their own shortcomings as mentioned in Section~\ref{sec:background}.
	Additionally, LPA relies heavily on the smoothness assumption that adjacent nodes tend to share similar labels.
	In contrast, we propose a novel sampling technique that allows parametric GNN models to learn interrelationships between labels by taking label information as input. The advantages of our method are as follows:
	\begin{itemize}[topsep=3pt,leftmargin=10pt]
	\item Capable of propagating features and labels during \emph{both training and inference stages}.
	\item Does not explicitly rely on the smoothness assumption of LPA, and can be conveniently adapted to various GNN architectures capable of handling heterogeneous and heterophily graphs where this assumption may break down \citep{zhu2020beyond,busbridge2019relational,PeiWCLY20,schlichtkrull2018modeling,yang2021graph}.
	\item Can be trained end-to-end, while avoiding the model learning trivial degenerate solutions, i.e., an identity mapping whereby the ground-truth labels merely pass directly from input to output training nodes.
	\end{itemize}
	
	Our method starts with a random split of $\mathcal{D}_{train}$ into several sub-datasets.
	For simplicity, we consider the case of two sub-datasets here, denoted as $\mathcal{D}^L_{train}$ and $\mathcal{D}^U_{train}$, respectively.
	Next, we set to zero the labels of $\mathcal{D}^U_{train}$, and learn to predict their original values.
	Specifically, the input for $\mathcal{D}^L_{train}$ contains both features and labels, while the input for $\mathcal{D}^U_{train}$ contains only features, where the labels used as inputs are set to zero-valued null vectors.
	During the final inference procedure, all labels in the training set are used as inputs to the model.
	We summarize this training procedure in Algorithm~\ref{algo:framework}, where $f_{\btheta}$ denotes an arbitrary GNN model with parameters $\btheta$; for further analysis of this label trick, please see \cite{wang2021why}.
	
	\begin{algorithm}[t]
		\caption{Label Usage for Graph Neural Networks}
		\label{algo:framework}
		\begin{algorithmic}[1]
			\REQUIRE $G$, $\bX$, $\bY$, the recycling times $R$
			\FOR{\textbf{\textnormal{each}} epoch}
			\STATE Obtain $\mathcal{D}^L_{train},\mathcal{D}^U_{train}$ by randomly splitting $\mathcal{D}_{train}$
			\STATE $\by^L_{i} \leftarrow \begin{cases}\by_i, & (v_i,\bx_i,\by_i)\in \mathcal{D}^L_{train}\\\mathbf 0, & \text{otherwise}\\\end{cases}$
			\label{line:linelabel}
			\STATE $\hat{\bY}^{(0)} \leftarrow f_{\btheta}(\bX \parallel \bY^L, \bA)$
			\FOR{$k\leftarrow1$ to $R$}
			\label{line:start}
			\STATE $\by_{i}^{(k-1)} \leftarrow \begin{cases}\by_i, & (v_i,\bx_i,\by_i)\in \mathcal{D}^L_{train}\\ \hat{\by}_{i}^{(k-1)}, & \text{otherwise}\\\end{cases}$
			\STATE $\hat{\bY}^{(k)}\leftarrow f_{\btheta}(\bX \parallel \bY^{(k-1)}, \bA)$
			\ENDFOR
			\label{line:end}
			\STATE Compute $\mathcal{L}(\btheta)$ and update $\btheta$ via back propagation.
			\ENDFOR
		\end{algorithmic}
	\end{algorithm}

	\minisection{Augmentation with Label Reuse}
	We further propose \textit{label reuse}, which recycles the predicted soft labels of the previous iteration and uses them labels as input.
	In this case, the labels of $\mathcal{D}_{train}^U$ and all test nodes are not assigned with zero-valued null vectors but the predicted results of the previous iteration.
	In Algorithm~\ref{algo:framework}, line~\ref{line:start} to line~\ref{line:end} presents the label reuse procedure.

	
	\begin{table*}[htbp]
	\hspace{-0.07in}\begin{minipage}[ht]{0.33\linewidth}
		\vskip 0.1in
		\begin{center}
			\resizebox{1.08\linewidth}{!}{
				\renewcommand{\arraystretch}{1.1}
				\begin{tabular}{ccc}
				\toprule
					\Large{\textbf{Loss}} & \Large{$\rho(z)$} & \Large{$\rho(\phi_{logit}(v))$}
					\rule{0pt}{2ex}\\
					\midrule
					\rule{0pt}{3ex}
					\Large{Logistic} & $z$ & $\log(1+\exp(-v))$ \\
					\rule{0pt}{4ex}
					\Large{Exponential} & $\exp(z)-1$ & $\exp(-v)$ \\
					\rule{0pt}{4ex}
					\Large{Sigmoid} & $1-\exp(-z)$& $\dfrac{1}{1+\exp(v)}$  \\
					\rule{0pt}{4ex}
					\Large{Savage} & $(1-\exp(-z))^2$ & $\dfrac{1}{(1+\exp(v))^2}$  \\
					\rule{0pt}{4ex}
					$\mathcal L_q$ & $\frac{1}{q}(1-\exp(-qz))$ & $\frac{1}{q}\left(1-\dfrac{1}{(1+\exp(-v))^{q}}\right)$  \\
					\rule{0pt}{4ex}
					\Large{\lossname} & $\log(\epsilon+z)-\log\epsilon$ & $\log(\epsilon+\log(1+\exp(-v)))-\log\epsilon$ \\
					\bottomrule
				\end{tabular}
			}
		\vskip 0.25in
		\captionof{table}{Loss functions with different $\rho(\cdot)$.}
		\label{table:loss}
		\end{center}
	\end{minipage}
	\hfill
    \begin{minipage}[ht]{0.666\linewidth}
		\begin{tabular}{cc}
		\includegraphics[scale=0.4]{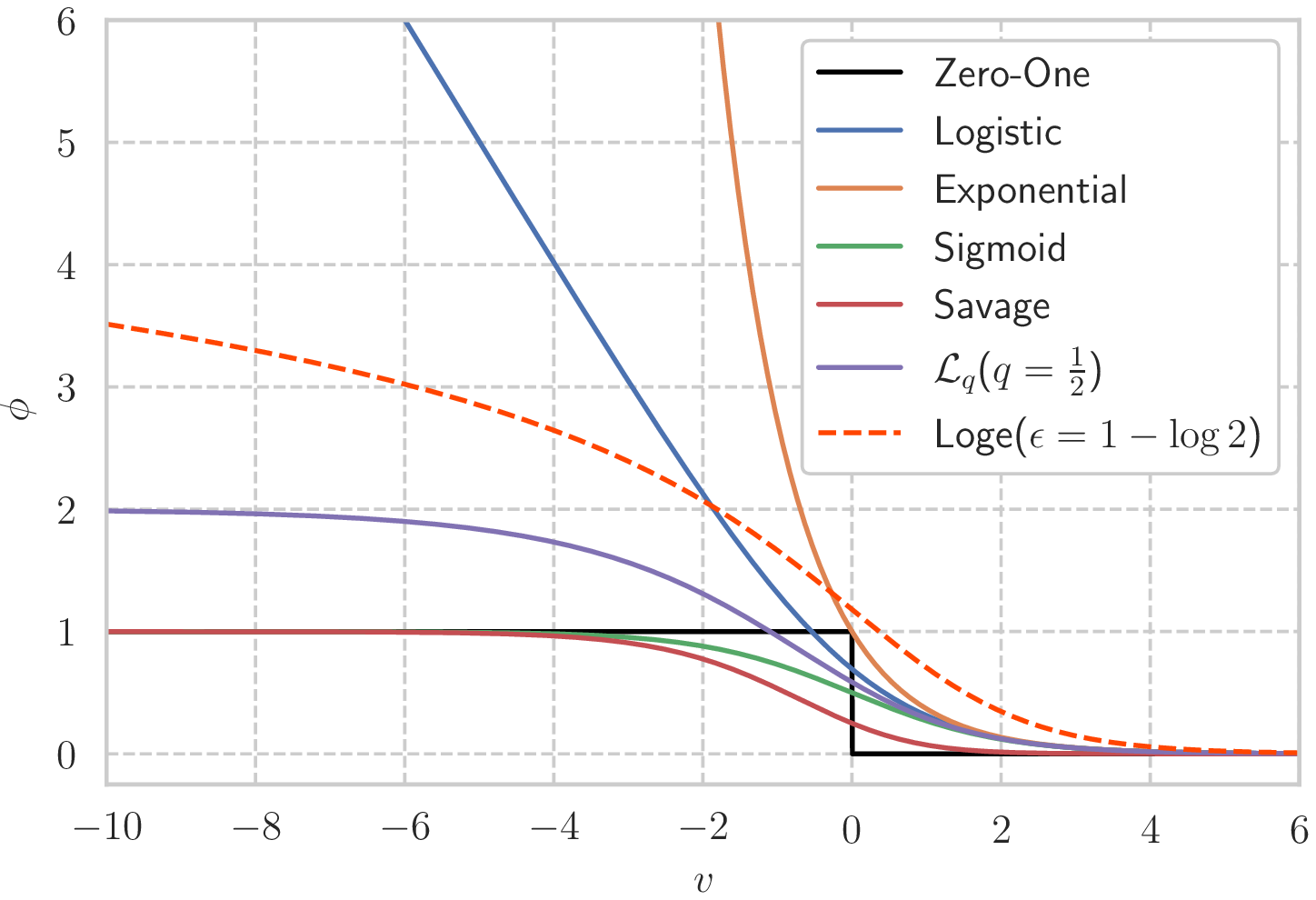}
		\includegraphics[scale=0.4]{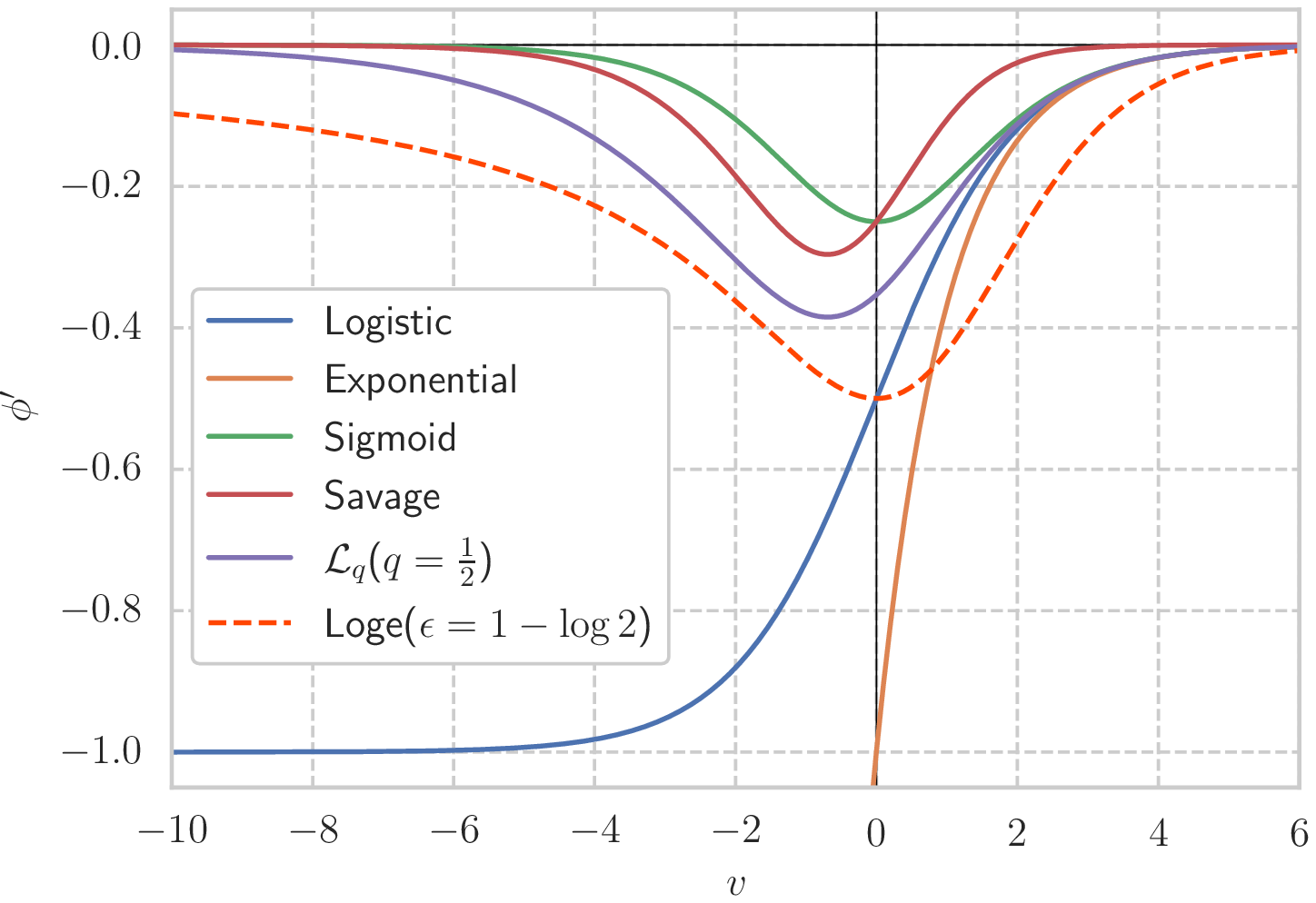}
		\end{tabular}
		\vskip -0.16in
		\captionsetup{width=.85\linewidth}
		\captionof{figure}{Visualization of various margin-based losses $\phi$ (\textit{left}) and their corresponding derivatives $\phi'$ (\textit{right}).
        }
		\label{fig:loss}
	\end{minipage}
    \end{table*}	
	
	\subsection{Robust Loss Function for Classification}
	In binary classification scenarios, given feature space $\mathcal X$ and label space $\mathcal \bY=\{-1,+1\}$, we aim to learn a classifier $g$ that maps $\bx\in\mathcal X$ to $\mathcal \bY$.
	The classifier follows the decision rule $g(\bx)=\text{sign}(f(\bx))$ for some mapping $f$ from $\mathcal X$ to $\mathbb R$.
	The optimization objective is to minimize the risk, defined as
	\begin{equation}
	    R_{\phi}(f):=\mathbb E_{\mathcal D}[\ell(f(\bx),y)]=\mathbb E_{\mathcal D}[\phi(yf(\bx))],
	\end{equation}
	where $\ell(f(\bx),y)$ is the loss function and $\phi:\mathbb R\rightarrow\mathbb R^+$ is known as the \textit{margin-based loss function}.  In this setting, choosing the loss function corresponds to choosing $\phi(\cdot)$.
	
	A straightforward choice for $\phi(\cdot)$ is the \textit{0-1 loss}
	\begin{equation}
	\ell_{0/1}(f(\bx),y)= \phi_{0/1}(yf(\bx)) \coloneqq H(-yf(\bx)),
	\end{equation}
	where $H(\cdot)$ denotes the Heaviside step function.
	However, $\phi_{0/1}(\cdot)$ is a discontinuous function and is therefore computationally challenging to optimize.
	As a result, instead of directly optimizing the \textit{0-1 loss}, we turn to using $\phi(\cdot)$ as an upper bound of $\phi_{0/1}(\cdot)$, often referred to as the \textit{calibrated surrogate loss}, for the optimization objective.
	
	Being the most commonly used loss function for classification, the \textit{logistic loss}, denoted as $\phi_{logit}(\cdot)$, provides a convex upper bound for $\phi_{0/1}(\cdot)$, which takes the form
	\begin{equation}
		\phi_{logit}(v)=\log(1+\exp(-v)).
	\end{equation}
    While the logistic loss performs satisfactorily in most cases, it suffers from sensitivity to outliers, whereas non-convex loss functions could be more robust \citep{masnadi2008design}. Motivated by this, we consider weakening the convexity condition, and thereby designing a quasi-convex loss to contribute robustness:
	\begin{equation}
		\phi_{\rho-logit}(v):=\rho(\phi_{logit}(v)),
	\end{equation}
	where $\rho:\mathbb R^+\rightarrow\mathbb R^+$ is a non-decreasing function.
	
	Some loss functions have been proposed to achieve outlier robustness, e.g., the \textit{Savage loss} \citep{masnadi2008design} and $\mathcal L_q$ loss \citep{zhang2018generalized}, which can be interpreted as choosing a suitable $\rho(\cdot)$.  Table~\ref{table:loss} summarizes different $\rho(\cdot)$ of these and other loss functions discussed.
	
	Here, we propose the \textit{\lossname loss} as a more preferable possibility for $\rho(\cdot)$:
	\begin{equation}
	    \rho_{\lossshortname}(z):=\log(\epsilon+z)-\log\epsilon,
	\end{equation}
	where $\epsilon$ is a tunable parameter and is fixed to $1-\log 2$ throughout this paper so that $\left.\frac{d^2}{dv^2}\phi_{\lossshortname}(v)\right|_{v=0}=0$ with $\phi_{\lossshortname}(v) := \rho_{\lossshortname}(\phi_{logit}(v))$.
	This implies that the derivative of the \lossname loss reaches its maximum magnitude at $z=0$, which is exactly the decision boundary for classification tasks.
	Since the derivative of $\phi(\cdot)$ is the weight of the data sample \citep{leistner2009robustness}, this may facilitate the optimization of classification accuracy.
	
	Figure~\ref{fig:loss} illustrates the visualization of \lossname loss and other losses along with their derivatives.
	As can be seen, our \lossname loss meets the following criteria:
	\begin{itemize}[topsep=3pt,leftmargin=10pt]
	\item While it helps to prevent outliers from dominating the training loss, it is nonetheless unbounded in such a way that the derivative converges slowly to $0$ as $v$ decreases, meaning that it still provides non-negligible gradient signals for the misclassified samples as desired.
	\item The maximum gradient magnitude occurs at $v=0$, which enhances the gradient signal near the decision boundary. The only other loss function with an analogously maximal gradient at $0$ is the Sigmoid loss; however, its tail converges to $0$ very fast, which can lead to a vanishing gradient problem.
	\item For correctly classified samples (i.e., $v>0$), the derivative converges to $0$ relatively quickly as with other loss functions; however, in this regime the gradient signal is less critical.
	\end{itemize}
	

	The \lossname loss can also be extended for multi-class classification tasks.  For this purpose, we formulate the labels in a one-hot fashion, where both $\by$ and $\hat \by$ are one-hot vectors, $\hat y_i$ denotes the value of the $i$-th element in $\hat \by$, and the predicted value of the target class is denoted by $\hat y_{class}$, meaning that the subscript \textit{class} refers to the index of the nonzero element of $\by$.
	The \lossname loss can then be formulated as
	\begin{equation}
		\ell_{\lossshortname}(\hat \by,\by)=\log\left(\epsilon-\log\frac{\exp({\hat y}_{class})}{\sum_{i=1}^C\exp({\hat y_i})}\right)-\log\epsilon.
	\end{equation}
	
	\subsection{Tweaking the GAT Architecture}
	\minisection{GAT with Symmetric Normalized Adjacency Matrix}
	We find the symmetric normalized adjacency matrix in GCN improves the performance at times, and yet GAT is not a natural extension of GCN. In order to better connect GAT with GCN, we first define the unnormalized attention matrix $\bA_{att}=\bD\balpha$, with $\balpha$ described in Eq.~(\ref{eqn:gat}).  Then the message passing rule with self-loops becomes
	\begin{equation}
		\bX^{(l+1)}=\sigma\left(\tilde\bD^{-\frac12}\tilde\bA_{att}\tilde\bD^{-\frac12}\bX^{(l)}\bW_0^{(l)}+\bX^{(l)}\bW_1^{(l)}\right),
	\end{equation}
	where $\tilde\bA_{att}=\bI+\bA_{att}$. 
	Note that when $\bA_{att}=\bA$, this variant is equivalent to Eq.~(\ref{eqn:trickgcn}) of GCN.
	
	
	\minisection{Other GAT Variants}
	The attention mechanism of the original GAT is described in Eq.~(\ref{eqn:gat}).  By replacing $\ba^T\bW$ with $\ba^T$, the computation of attention value in Eq.~(\ref{eqn:gat}) can be simplified to
	\begin{equation}
		\alpha_{ij}=\frac{\exp\left(\mathrm{LeakyReLU}\left(\ba^T[\bx_i\parallel \bx_j]\right)\right)}{\sum_{r\in\mathcal{N}(v_i)}\exp\left(\mathrm{LeakyReLU}\left(\ba^T[\bx_i\parallel \bx_r]\right)\right)},
	\end{equation}
	where for simplicity we henceforth omit the layer-wise superscripts.
    Another variant is the non-interactive GAT, which performs similarly to and at times better than the original form, and can be expressed as
	\begin{equation}
		\alpha_{ij}=\frac{\exp\left(\mathrm{LeakyReLU}\left(\ba^T \bx_j\right)\right)}{\sum_{r\in\mathcal{N}(v_i)}\exp\left(\mathrm{LeakyReLU}\left(\ba^T \bx_r\right)\right)}.
	\end{equation}
    We also propose a GAT variant that exploits the edge features in the graph:
	\begin{equation}
		\alpha_{ij}=\frac{\exp\left(\mathrm{LeakyReLU}\left(\ba^T[\bx_i^V\parallel\bx_j^V\parallel \bx_{ij}^E]\right)\right)}{\sum_{r\in\mathcal{N}(v_i)}\exp\left(\mathrm{LeakyReLU}\left(\ba^T[\bx_i^V\parallel \bx_r^V\parallel \bx_{ij}^E]\right)\right)},
	\end{equation}
	where $\bx^V$ and $\bx^E$ denote node and edge features respectively. The time complexity of computing one layer of a single-headed GAT with $C^V$ node features, $C^E$ edge features and $F$ filters is $O(|V|C^VF+|E|C^E+|E|F)$.

	\section{Experiments}
	In this section, we examine the performance of each method through ablation experiments, reporting the mean classification accuracy for multi-class classification tasks and Area Under the ROC Curve (ROC-AUC) for binary classification tasks. We choose three commonly used citation network datasets, Cora, Citeseer and Pubmed \citep{DBLP:journals/aim/SenNBGGE08}, and three relatively large datasets from OGB \citep{hu2020open}, ogbn-arxiv, ogbn-proteins and ogbn-products, as well as Reddit, a dataset of posts from the Reddit website.\footnote{\url{https://snap.stanford.edu/graphsage/}.}
	Some statistics of these datasets are presented in Table~\ref{table:dataset-statistics}.
	Since ogbn-proteins is a dataset with edge features and ogbn-products is a huge dataset, we adopt neighbor sampling for them due to memory constraints.
	For Cora, Pubmed and Citeseer, we report the average scores and standard deviations after 100 runs, and for the relatively larger datasets ogbn-arxiv, ogbn-proteins, ogbn-products and Reddit, we report mean scores and standard deviations after 10 runs.
	All experiments were implemented using the Deep Graph Library (DGL) \citep{wang2019deep}.\footnote{Reproducible code based on DGL with instructions is available at \burl{https://github.com/espylapiza/Bag-of-Tricks-for-Node-Classification-with-Graph-Neural-Networks}.} 
	
	\begin{table}[t]
		\caption{Datasets statistics, where \textit{label rate} denotes the proportion of labeled nodes used for training to the total nodes.}
		\label{table:dataset-statistics}
		\begin{center} 
			\resizebox{0.98\linewidth}{!}{
				\begin{tabular}{lrrcc}
					\toprule
					\textbf{Dataset}  & \textbf{\#Nodes} & \textbf{\#Edges} & \textbf{Metric} & \textbf{Label rate} \\
					\midrule
					Cora              & 2,708           & 5,429           & Accuracy        & 5.2\% \\
					Citeseer          & 3,327           & 4,732           & Accuracy        & 3.6\% \\
					Pubmed            & 1,9717          & 44,338          & Accuracy        & 0.03\% \\
					Reddit            & 232,965         & 114,615,892     & Accuracy        & 65.9\% \\
					Arxiv             & 169,343         & 1,166,243       & Accuracy        & 53.7\% \\
					Proteins          & 132,534	        & 39,561,252      & ROC-AUC         & 65.4\% \\
					Products          & 2,449,029       & 61,859,140      & Accuracy        & 8.0\% \\
					\bottomrule
				\end{tabular} 
			}
		\end{center}
	\end{table}
	
	\begin{table}[t]
		\caption{Accuracy results (as measured by classification accuracy and ROC-AUC for ogbn-arxiv and ogbn-proteins, respectively) of different datasets and models in terms of label usage and GAT variant. \textit{GCN+linear} indicates the GCN variant with a residual connection of Eq.~(\ref{eqn:trickgcn}). \textit{GAT*} indicates the GAT variant that incorporates the edge features.}
		\label{table:result-label-usage}
		\begin{center}
			\resizebox{0.98\linewidth}{!}{
				\begin{tabular}{llcc}
					\toprule
					\textbf{Dataset} & \textbf{Model}    & \textbf{Label Usage} & \textbf{Accuracy}(\%)  \\
					\midrule
					Arxiv & GCN & --                & 72.48 ± 0.11    \\
					Arxiv & GCN & label as input    & 72.64 ± 0.10    \\
					Arxiv & GCN & label reuse       & \textbf{72.78 ± 0.17}    \\
					\midrule
					Arxiv & GCN+linear              & --                & 72.74 ± 0.13    \\
					Arxiv & GCN+linear              & label as input    & 73.13 ± 0.14    \\
					Arxiv & GCN+linear              & label reuse       & \textbf{73.22 ± 0.13}    \\
					\midrule
					Arxiv & GAT                     & --                & 73.20 ± 0.16    \\
					Arxiv & GAT                     & label as input    & 73.24 ± 0.10    \\
					Arxiv & GAT                     & label reuse       & \textbf{73.43 ± 0.13}    \\
					\midrule
					Arxiv & GAT(norm.adj.)          & --                  & 73.59 ± 0.14    \\
					Arxiv & GAT(norm.adj.)          & label as input      & 73.66 ± 0.11    \\
					Arxiv & GAT(norm.adj.)          & label reuse         & 73.91 ± 0.12    \\
					Arxiv & GAT(norm.adj.)          & label reuse+C\&S    & \textbf{73.95 ± 0.12}    \\
					\midrule
					Arxiv & AGDN                    & --                  & 73.75 ± 0.21    \\
					Arxiv & AGDN                    & label as input      & \textbf{73.98 ± 0.09}    \\
					\midrule
					Proteins & GCN                  & --                  & 80.07 ± 0.95    \\
					Proteins & GCN                  & label as input      & \textbf{80.80 ± 0.56}    \\
					\midrule
					Proteins & GAT*                 & --                  & 87.47 ± 0.16    \\
					Proteins & GAT*                 & label as input      & \textbf{87.65 ± 0.08}    \\
					\bottomrule
				\end{tabular}
			}
		\end{center}
	\end{table}
	
	\minisection{Label Usage}
	There are two principal factors that determine the benefits of using labels as inputs during training. One is the proportion of graph nodes with labels available for training, and the other is the training accuracy. We investigate the performance of label as input and label reuse on datasets with a relatively large proportion of training set and low training accuracy. The results are reported in Table~\ref{table:result-label-usage}. Here our approach improves the performance consistently with only a small increase in parameters. Furthermore, we can further improve the performance by combining our method with C\&S \citep{huang2020combining}.
	
	\minisection{Loss Functions}
	We evaluate the performance of our loss function on datasets with classification accuracy as the metric. Results are reported in Table~\ref{table:result-loss-function}, where each model is trained with the same hyperparameters, varying only the loss functions.
	As shown, while the robust Savage loss performs well on some small datasets, it performs considerably worse on larger datasets.
	Meanwhile, the \lossname loss outperforms other losses on most datasets.
	
	\begin{table}[t]
		\caption{Comparative results of loss functions on different datasets and models, where $\epsilon$ of the \lossname loss is $1-\log 2$.}
		\label{table:result-loss-function}
		\begin{center}
			\resizebox{1\linewidth}{!}{
				\begin{tabular}{llccc}
					\toprule
					\textbf{Dataset} & \textbf{Model} & \multicolumn{3}{c}{\textbf{Accuracy}(\%)} \\
					&                & \textbf{Logistic} & \textbf{Savage} & \textbf{\lossname} \\
					\midrule
					Cora        & MLP            & 59.72 ± 1.01 & \textbf{61.10 ± 0.91} & 60.39 ± 0.74 \\
					Cora        & GCN            & 82.26 ± 0.84 & 81.65 ± 0.74 & \textbf{82.60 ± 0.83} \\
					\midrule
					Citeseer    & MLP            & 57.75 ± 1.05 & \textbf{59.60 ± 0.92} & 59.07 ± 0.98 \\
					Citeseer    & GCN            & 71.13 ± 1.12 & 71.10 ± 1.22 & \textbf{72.49 ± 1.12} \\
					\midrule
					Pubmed      & MLP            & 73.15 ± 0.68 & \textbf{73.39 ± 0.62} & 72.93 ± 0.65 \\
					Pubmed      & GCN            & 78.89 ± 0.71 & 78.91 ± 0.63 & \textbf{78.93 ± 0.69} \\
					\midrule
					Reddit      & MLP            & 72.98 ± 0.09 & 68.64 ± 0.29 & \textbf{73.12 ± 0.09}  \\
					Reddit      & GCN            & \textbf{95.22 ± 0.04} & 92.29 ± 0.48 & 95.18 ± 0.03  \\
					\midrule
					Arxiv       & MLP            & 56.18 ± 0.14 & 51.97 ± 0.20 & \textbf{56.72 ± 0.15}  \\
					Arxiv       & GCN            & 71.77 ± 0.34 & 68.47 ± 0.32 & \textbf{72.43 ± 0.16}  \\
					Arxiv       & GAT            & 73.08 ± 0.26 & 69.58 ± 1.00 & \textbf{73.20 ± 0.16}  \\
					Arxiv       & GAT(norm.adj.) & 73.29 ± 0.17 & 69.22 ± 1.48 & \textbf{73.59 ± 0.14}  \\
					\midrule
					Products    & MLP            & 62.90 ± 0.16 & 58.13 ± 1.03 & \textbf{63.20 ± 0.13}  \\
					Products    & GAT            & 80.99 ± 0.16 & 77.48 ± 0.14 & \textbf{81.39 ± 0.14}  \\
					\bottomrule
				\end{tabular}
			}
		\end{center}
	\end{table}
	
	\begin{table}[t]
	    \caption{Results of GAT variant. GAT+norm.adj. corresponds to GAT with symmetric normalized adjacency.}
		\label{table:result-gcn-gat-variant}
		\begin{center} 
			\resizebox{0.66\linewidth}{!}{
				\begin{tabular}{lcccc}
					\toprule
					\textbf{Dataset}  &  \multicolumn{2}{c}{\textbf{Accuracy}(\%)}  \\
					&   \textbf{vanilla GAT} & \textbf{GAT+norm.adj.}  \\
					\midrule
					Cora              & 83.41 ± 0.74 & \textbf{83.72 ± 0.74}  \\
					Citeseer          & 71.92 ± 0.92 & \textbf{72.25 ± 1.04}  \\
					Pubmed            & 78.43 ± 0.64 & \textbf{78.77 ± 0.54}  \\
					Reddit            & 96.97 ± 0.04 & \textbf{97.06 ± 0.05}  \\
					Arxiv             & 73.20 ± 0.16 & \textbf{73.59 ± 0.14}  \\
					\bottomrule
				\end{tabular} 
			}
		\end{center}
	\end{table}
	
	\minisection{GAT Variants}
	To explore the effect of the symmetric normalized adjacency matrix on GAT, we compare its performance with the original GAT on 5 datasets. The results are reported in Table~\ref{table:result-gcn-gat-variant}. We see that GAT with a normalized adjacency matrix achieves higher performance on all datasets. Nevertheless, we recommend choosing the appropriate adjacency matrix for different datasets.
	In Table~\ref{table:result-label-usage}, our GAT variant that incorporates the edge features \textit{outperforms all prior methods applied to the ogbn-proteins dataset by a significant margin at the time of our post to the OGB leaderboard.}

	\section{Conclusion}
	In this paper, we present a new framework for combining feature and label propagation, propose a robust loss function, and investigate several tricks for training deep GNNs with promising performance.
	These techniques can be applied to various GNN models, which generally only require minor modifications to the data processing, loss function, or architecture.
	
	\section{Acknowledgements}
	We thank the support of National Natural Science Foundation of China (Grant No. 61702327, 61772333, 61632017) and Wu Wen Jun Honorary Doctoral Scholarship, AI Institute, Shanghai Jiao Tong University.


\bibliographystyle{ACM-Reference-Format}
\balance
\bibliography{main}

\end{document}